\documentclass[
]{ceurart}

\usepackage{listings}
\usepackage{tikz}
\usetikzlibrary{positioning,shapes.geometric} 
\usepackage{csquotes}
\usepackage{orcidlink}

\begin{document}

\copyrightyear{2025}
\copyrightclause{Copyright for this paper by its authors. Use permitted under Creative Commons License Attribution 4.0 International (CC BY 4.0).}

\conference{The Second International Workshop on Hypermedia Multi-Agent Systems (HyperAgents 2025), in conjunction with the 28th European Conference on Artificial Intelligence (ECAI 2025); October 26, 2025, Bologna, Italy}

\title{Affordance Representation and Recognition for Autonomous Agents}



\author[1]{Habtom Kahsay Gidey}[%
orcid=0000-0001-5802-2606,
email=habtom.gidey@tum.de,
]
\cormark[1]
\address[1]{Technische Universität München, München, Germany}
\address[2]{Jessy Works, München, Germany}

\author[2]{Niklas Huber}[
email=niklas.huber@jessyworks.com,
]

\author[1]{Alexander Lenz}[%
email=alex.lenz@tum.de,
]

\author[1]{Alois Knoll}[%
orcid=0000-0003-4840-076X,
email=knoll@tum.de,
]

\cortext[1]{Corresponding author.}

\begin{abstract}
The autonomy of software agents is fundamentally dependent on their ability to construct an actionable internal world model from the structured data that defines their digital environment, such as the Document Object Model (DOM) of web pages and the semantic descriptions of web services. However, constructing this world model from raw structured data presents two critical challenges: the verbosity of raw HTML makes it computationally intractable for direct use by foundation models, while the static nature of hardcoded API integrations prevents agents from adapting to evolving services.

This paper introduces a pattern language for world modeling from structured data, presenting two complementary architectural patterns. The~\textit{DOM Transduction Pattern} addresses the challenge of web page complexity by~\textit{distilling} a verbose, raw DOM into a compact, task-relevant representation or world model optimized for an agent's reasoning core. Concurrently, the~\textit{Hypermedia Affordances Recognition Pattern} enables the agent to dynamically enrich its world model by parsing standardized semantic descriptions to discover and integrate the capabilities of unknown web services at runtime. Together, these patterns provide a robust framework for engineering agents that can efficiently construct and maintain an accurate world model, enabling scalable, adaptive, and interoperable automation across the web and its extended resources.
\end{abstract}

\begin{keywords}
  Autonomous Agents \sep
  Affordances \sep
  World Modeling \sep
  Design Patterns \sep
  Web Automation.
\end{keywords}

\maketitle
\section{Introduction}

Recent advances and the abundance of foundation models are driving a new generation of software agents performing complex cognitive tasks in dynamic, mixed-reality ecosystems~\cite{gur2023webagent,ning2025survey,macedo2024evolving}.
For these agents to operate effectively, they must perceive and understand their environment to build an internal~\textit{world model}, an actionable representation that guides their reasoning and planning~\cite{ha2018world,lecun2022path,richens2025worldmodels}. 
Although significant research has focused on visual perception from pixels, many digital environments are built upon rich, structured data sources like the hypermedia of the Document Object Model (DOM) of web pages and the descriptive interfaces of web services~\cite{yang2024agentoccam,he2024webvoyager}. 
To this end, leveraging this explicit structure offers a more efficient and deterministic path to building a world model~\cite{tan2025htmlrag,liu2025wepo}.

However, directly consuming this structured data presents two critical architectural challenges that hinder the development of truly autonomous agents~\cite{assouel2023unsolved,ning2025survey}. 
First, the verbosity of raw HTML is a major bottleneck for agents using foundation models for reasoning and planning~\cite{gur2023webagent}. 
Empirical studies of DOM pruning show that the majority, often 80 - 90\%, of tokens in raw HTML consist of non-semantic markup such as scripts, styles, and trackers~\cite{tan2025htmlrag,yang2024agentoccam}, 
which can overwhelm the model's, i.e., LLMs, context window, degrade reasoning quality, and incur high computational costs~\cite{lu2025buildweb}. 
This is the fundamental~\textit{challenge of representation}, where the raw state of the world is too complex for the agent's cognitive core to process efficiently~\cite{liu2025wepo,kim2025nexteval}.

Second, agents must interact with a dynamic ecosystem of services and devices in the increasingly interconnected Web of Things (WoT) and microservice architectures~\cite{gandon2021merry,castellucci12025towards,gidey2022document}. 
In such cases, the traditional approach of hardcoding binary interfaces, such as API endpoints and interaction logic, creates brittle, tightly coupled systems that cannot adapt when a service changes or a new device is introduced in the web microcosm~\cite{gidey2023towards,assouel2023unsolved}. 
This is the~\textit{challenge of interoperability, adaptability, and discovery}, i.e., recognition of affordances, where the agent's world model is static and cannot be updated to reflect a changing digital environment~\cite{lu2025buildweb,liu2025wepo}.

This paper introduces preliminary work on a pattern language~\cite{gidey2017grounded} that addresses these challenges by framing them as problems of world model construction~\cite{ha2018world,lecun2022path,richens2025worldmodels}. We present two architectural design patterns that provide reusable solutions for building and enriching an agent's world model from structured data:

\begin{enumerate}
\item \textbf{The DOM Transduction Pattern:} A pattern for distilling a complex, raw DOM into a compact, task-relevant world model optimized for an agent's reasoning core. 
\item \textbf{The Hypermedia Affordances Recognition Pattern:} A pattern for dynamically enriching the world model by parsing standardized semantic descriptions of web services to discover and integrate their capabilities at runtime. 
\end{enumerate}

Combined with other percepts, these patterns provide a robust framework for engineering agents that can efficiently and adaptively interact with the structured web and its connected extended resources and the web of things~\cite{castellucci12025towards,lecun2022path}.

\section{Background}
This research work covers several domains; in particular, it is situated at the intersection of hypermedia multi-agent systems, world modeling, and cognitive automation~\cite{gidey2023user,gandon2021merry,ha2018world,castellucci12025towards,lecun2022path,richens2025worldmodels}.

\subsection{Web Agents Observation Space} 

The challenge of processing verbose HTML for LLM-based agents has become a significant area of research~\cite{assouel2023unsolved,ning2025survey}. 
The core problem is that the agent's observation space, when represented by a raw DOM, is misaligned with the LLM's processing capabilities. 
As a result, this has led to the development of various techniques for DOM cleaning and simplification~\cite{tan2025htmlrag,yang2024agentoccam,liu2025wepo}. 
Recent work on LLM-based web agents such as WebVoyager, Agent-E, and AgentOccam has highlighted the critical importance of managing the agent's observation space~\cite{abuelsaad2024agent, he2024webvoyager,yang2024agentoccam}. 
The primary challenge lies in handling complex and verbose HTML, which motivates approaches such as DOM distillation and HTML pruning. 
Several approaches have emerged, collectively known as DOM distillation or HTML pruning, which aim to simplify the DOM to make it more tractable for LLMs. 
For example, AgentOccam focuses on refining the observation space to better align with the LLM's capabilities~\cite{yang2024agentoccam}. 
More recent work, such as HtmlRAG, has introduced block-tree-based pruning methods and concrete methods to clean and compress HTML while preserving its structure~\cite{tan2025htmlrag}. 
This concept of webpage segmentation has a long history in information extraction, aiming to break down a page into semantically related parts to improve downstream tasks~\cite{abuelsaad2024agent,he2024webvoyager,yang2024agentoccam}. 
These approaches demonstrate the critical importance of preprocessing the DOM. Our DOM Transduction Pattern then aims to formalize these emerging best practices into a reusable architectural solution.

\subsection{Hypermedia Multi-agent Systems and World Models via Hypermedia}
 
The second challenge, interoperability, is addressed by principles from hypermedia multi-agent systems~\cite{gandon2021merry}. 
The core idea is~\textit{Hypermedia as the Engine of Application State (HATEOAS)}, a fundamental constraint of the REST architectural style~\cite{kelly2023hal}. 
HATEOAS states that a client should navigate an application entirely through links provided dynamically by the server, eliminating the need for a developer in the middle or hardcoded endpoints and thus decoupling the client from the server. 
While the adoption of HATEOAS in general web APIs has been debated, its principles are ideally suited for autonomous agents that require dynamic discovery and adaptation of affordances.

The most concrete and standardized application of these principles is the W3C Web of Things (WoT) framework~\cite{lagally2023wot}. 
Its central component, the~\textit{Thing Description (TD)}, is a JSON-LD document that provides machine-readable~\enquote{capability knowledge} for any given~\enquote{Thing,} such as a device, service, or other web resource. 
A TD specifies a resource’s metadata and its~\textit{Interaction Affordances}, the~\enquote*{Properties},~\enquote*{Actions}, and~\enquote*{Events} it exposes, along with the specific~\enquote*{Protocol Bindings} (e.g., HTTP, MQTT) required for interaction. 
By parsing a TD, an agent can dynamically learn what a service can do and how to communicate with it without prior, hardcoded knowledge. 
This is complemented by \textit{WoT Discovery} mechanisms, which define how agents can find relevant Thing Descriptions, for instance, through a searchable Thing Description Directory (TDD)~\cite{wot-discovery,gidey2022document}.

Our Hypermedia Affordances Recognition Pattern formalizes this HATEOAS-based discovery process as a key mechanism for enriching an agent’s world model at runtime.

These ideas connect to the notion of a cognitive map or a world model, grounded in the foundational work on world models~\cite{ha2018world,lecun2022path,richens2025worldmodels}.

\section{Pattern Language for Structured Perception}
To systematically address the challenges of building world models from structured data, here we adopt the established software engineering methodology, i.e., design patterns~\cite{beck1987using,gidey2019modeling,gidey2017grounded}. 
A pattern is a reusable design or architectural solution to a recurring problem within a given context, forming a~\enquote{pattern language} that captures and communicates expert architectural knowledge. 
By applying this methodology, we can systematically investigate and codify solutions for agents' recurring perceptual challenges when interacting with structured digital environments.

Consequently, to ensure a rigorous description of each pattern, we employ a comprehensive cataloging template adapted from standard pattern documentation formats. This template organizes each pattern into three main sections: the~\textit{Problem Space}, which defines the context, problem, and motivating forces; the~\textit{Solution Space}, which details the solution's description, components, flow, and formal constraints; and~\textit{Application and Evaluation}, which discusses consequences and implementation. 
This format is deliberately preferred to facilitate future formalization and ensure the reproducibility of our proposed solutions~\cite{gidey2019modeling,gidey2018factum}.

\subsection{The DOM Transduction Pattern}

\subsubsection{Problem}
An LLM-based agent must understand and interact with a web page, but the raw HTML DOM is too verbose and noisy. 
It exceeds the LLM's context window, contains irrelevant information that degrades reasoning, and incurs high computational costs. 
To that end, the agent needs a simplified yet structurally coherent representation of page affordances to build its world model~\cite{assouel2023unsolved,ning2025survey}.

\subsubsection{Solution}
\begin{figure}[!htbp]
\centerline{\includegraphics[scale=0.56]{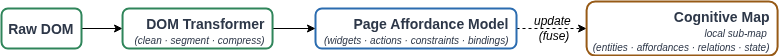}}
\caption{\textit{Flow and components of the DOM Transduction Pattern. Raw DOM is distilled into an affordance representation, the Page Affordance Model (PAM), and then fused with other structured percepts, such as service contracts and WoT Things, to update the Cognitive Map, the agent’s world model.}}
\label{fig:domTransductionPattern}
\end{figure}

This pattern introduces a~\textit{DOM Transformer} component within the agent's perception module. 
As illustrated in Fig.~\ref{fig:domTransductionPattern}, this component ingests raw DOM and applies a series of transformations to distill and represent it in a compact, task-relevant world model with affordances. 
This process typically involves the following.
\begin{enumerate}
    \item \textbf{Cleaning:} Removing universally irrelevant tags like~\enquote*{\texttt{<script>}} and~\enquote*{\texttt{<style>}}, which can substantially shrink the DOM size~~\cite{tan2025htmlrag,yang2024agentoccam}.
    \item \textbf{Pruning:} Intelligently removing content that is irrelevant to the current task, for example, block-tree-based pruning strategies or embedding-based relevance filtering~\cite{liu2025wepo,kim2025nexteval}. 
    \item \textbf{Compact Representation:} Converting cleaned HTML into token-efficient encodings such as Emmet notation~\cite{tan2025htmlrag}.
    \item \textbf{LLM-as-Transformer:} Using a smaller LLM to summarize the DOM before passing it to a larger reasoning model~\cite{assouel2023unsolved}.
\end{enumerate}

The output is a simplified DOM that preserves the essential structure needed for interaction while being optimized for LLM processing. 
Fig.~\ref{fig:domTransductionPattern} illustrates the flow and components of the perception pipeline for the~\textit{DOM Transduction Pattern}.

\subsubsection{Architectural Constraints}
\begin{enumerate}
\item The input must be a structured DOM tree.
\item The transformation is a unidirectional data flow from Raw DOM into a structured representation of the Page Affordance Model.
\item The output, Page Affordance Model, must preserve the essential structure of task-relevant interactive elements.
\item The DOM Transformer must be a decoupled perception component.
\end{enumerate}

\subsubsection{Application and Evaluation}
\begin{itemize}
\item \textbf{Consequences:}
\begin{itemize}
\item \textit{Benefits:} Enables automation on complex pages by overcoming context window limitations; significantly reduces cost and latency; improves agent reliability by providing a cleaner, more focused context.
\item \textit{Liabilities:} Designing a robust DOM Transformer is a non-trivial engineering task; an overly aggressive pruning strategy can lead to critical failures by removing necessary elements.
\end{itemize}

\item \textbf{Implementation Considerations:} 
\begin{itemize}
    \item \textit{Rule-Based Filtering:} Pattern-matching and rule-based parsing techniques can support the selective removal of predefined tags and attributes from the DOM.
    \item \textit{Block-Based Pruning:} Partitioning the DOM into semantic blocks, combined with task-aware relevance scoring, for example, embedding similarity to task descriptions, can provide effective strategies for discarding irrelevant content~\cite{tan2025htmlrag}.
    \item \textit{Compact Representation:} Structural encoding methods, such as token-efficient notations, can provide compressed forms of the cleaned DOM while preserving essential hierarchy and relationships~\cite{tan2025htmlrag}.
    \item \textit{LLM-as-Transformer:} Cascaded model strategies, where a smaller model distills or summarizes the DOM before forwarding it to a more capable reasoning model, can offer efficiency gains~\cite{assouel2023unsolved}.
\end{itemize}
\end{itemize}

\subsection{The Hypermedia Affordances Recognition Pattern}

\subsubsection{Problem}
An agent must operate in a dynamic ecosystem of distributed services and IoT devices (the~\enquote{Web of Things}). 
Hardcoding the API and having a developer in the middle for each service makes the agent brittle and unable to adapt to new or updated services. 
It must have perceptual skills to dynamically discover, recognize, and understand the capabilities, or affordances, of any web service or resource it encounters~\cite{gandon2021merry,gidey2023towards}.

\subsubsection{Solution}
This pattern, based on HATEOAS principles~\cite{kelly2023hal}, requires that services expose their capabilities through standardized, machine-readable semantic descriptions. 
The agent's perception module contains an~\textit{Affordance Parser} that fetches and interprets these descriptions. 
The canonical implementation is the~\textit{W3C WoT Thing Description (TD)}, a JSON-LD document that specifies two key elements~\cite{lagally2023wot}:
\begin{itemize}
    \item \textbf{Interaction Affordances:} The~\enquote*{Properties} (readable/writable state),~\enquote*{Actions} (invokable functions), and~\enquote*{Events} (subscribable notifications) the service offers.
    \item \textbf{Protocol Bindings:} These are the specific technical instructions that detail how an agent can interact with each of a service's capabilities or affordances.
\end{itemize}

\begin{figure}[!htbp]
\centerline{\includegraphics[scale=0.57]{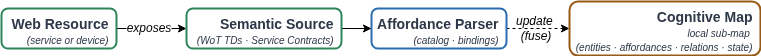}}
\caption{The flow of the Hypermedia Affordances Recognition Pattern involves the agent discovering a semantic description and parsing it into an affordances catalog which updates the Cognitive Map.}
\label{fig:hyperAffordancesRecognition}
\end{figure}

By parsing a TD, the agent dynamically learns how to interact with services at runtime.
As shown in Fig.~\ref{fig:hyperAffordancesRecognition}, this enables adaptability and robustness, allowing agents to autonomously integrate new devices and services without prior hardcoding.

\subsubsection{Architectural Constraints}
\begin{enumerate}
\item The agent and the resource it interacts with must be fully decoupled with no pre-configured, hardcoded dependencies on each other.
\item The resource must expose its capabilities and make them known by publishing a standardized semantic description.
\item All agent interactions must be driven by affordances discovered in the description. 
\item The resource's description dictates the interaction protocol and all communication specifics, which are determined by the resource itself, not the agent.
\end{enumerate}

\subsubsection{Application and Evaluation}
\begin{itemize}
\item \textbf{Consequences:}
\begin{itemize}
\item \textit{Benefits:} Enables extreme adaptability and robustness, allowing agents to autonomously integrate new devices and services on the fly; simplifies the development of large-scale, interoperable systems.
\item \textit{Liabilities:} Requires device and service providers to correctly implement and host a Thing Description, which can be a barrier to adoption; a poorly written TD can lead to agent errors.
\end{itemize}

\item \textbf{Implementation Considerations:} 
\begin{itemize}
\item Affordance parsing can be implemented using JSON-LD libraries and WoT toolkits. 
\item Agents must include client libraries for common protocols (HTTP, MQTT, etc.), and can use WoT Discovery mechanisms such as TD directories to find services~\cite{lagally2023wot}.
\end{itemize}
\end{itemize}

\section{Cognitive Map as a Unified World Model}

The two patterns embody two complementary modes of perception: one focused on distilling and representing a known, complex environment, and the other on discovering and integrating unknown entities. 
Jointly, they contribute to the construction of a unified cognitive map or a world model, as emphasized in the foundational work on world models~\cite{ha2018world,lecun2022path,richens2025worldmodels,gidey2023modeling}.

Their flow can be illustrated with a simplified practical example. An agent tasked with booking a hotel first applies the~\textit{DOM Transduction Pattern} to parse the booking website, producing a simplified world model of the form. 
Within this model, it discovers a hyperlink labeled~\textit{\enquote{Smart Room Controls}}. 
Following this link, the agent receives a W3C Thing Description for the room's environmental controls. 
It then switches to the~\textit{Hypermedia Affordances Recognition Pattern} to parse the description, dynamically enriching its world model with new capabilities, for example, discovering a~\textit{\enquote*{thermostat}} and a~\textit{\enquote*{setTemperature}} action. 
The agent can now not only complete the booking but also offer to pre-set the room temperature, an affordance discovered and integrated entirely at runtime.

\begin{figure}[!htbp]
\centerline{\includegraphics[scale=0.55]{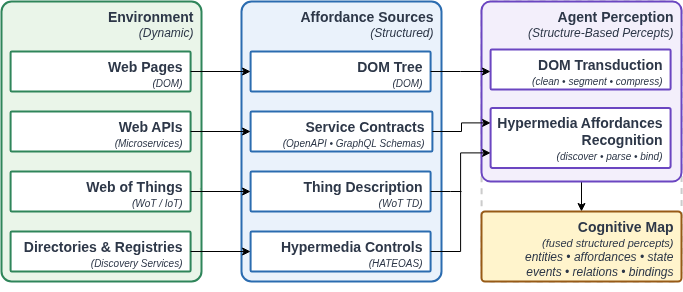}}
\caption{High-level flow from structured environments (web services, devices) to a unified cognitive map. Vision is excluded; only structured inputs are considered.}
\label{fig:cogMap}
\end{figure}

This composition of patterns is central to the agent's perception architecture, as shown in Fig.~\ref{fig:cogMap}. 
The diagram illustrates how different percepts and affordances, such as DOM trees, thing descriptions, and service contracts, are fused into a unified cognitive map.
The DOM Transduction Pattern processes HTML structures, while the Hypermedia Affordances Recognition Pattern interprets service descriptions. 
Concurrently, these complementary perceptual streams provide the agent with an adaptable and semantically rich representation of its environment in the evolving ecosystem.

\section{Conclusion and Outlook}
This paper introduced a preliminary pattern language to address key challenges in web and service interaction by presenting two architectural patterns that enable autonomous agents to construct and maintain an actionable world models from structured data. 
The DOM Transduction Pattern distills complex web pages into tractable affordance representations for LLM-based reasoning, while the Hypermedia Affordances Recognition Pattern enables dynamic discovery of service capabilities, ensuring interoperability and adaptability. 

The primary contribution of this work is a principled, reusable framework for engineering hypermedia multi-agent systems that can build and maintain accurate world models from the explicit structure of their environment. 
This enables the development of agents that are more efficient, scalable, resilient to change, and capable of predictive reasoning, compared to those relying on brittle, hardcoded logic.

As an outlook, this work represents one half of a larger vision. 
Our future research will extend these structure-based patterns with visual counterparts.
The long-term goal is a comprehensive pattern language for multi-modal perception, enabling agents to fuse structured and visual percepts. 
This will allow an agent, for example, to use the DOM Transduction pattern on a website but switch to visual parsing when structured representations are unavailable.
This ability to intelligently select and adapt perceptual modalities will advance the next generation of autonomous agents toward human-level competence in digital environments.

\section*{Declaration on Generative AI}
During the preparation of this work, the authors used Grammarly and ChatGPT in order to: Grammar and spelling check, Paraphrase and reword. After using this tool/service, the authors reviewed and edited the content as needed and take full responsibility for the publication’s content.

\bibliography{bibliography}

\appendix



\end{document}